\documentclass[runningheads]{llncs}
\usepackage[T1]{fontenc}
\usepackage{xcolor,soul}
\usepackage{orcidlink}
\usepackage{graphicx}
\usepackage{amsmath}
\usepackage{amssymb}
\usepackage{mathrsfs}
\usepackage{todonotes}
\usepackage{hyperref}
\usepackage[dvipsnames]{xcolor}
\begin{document}
\title{From Performance to Understanding: A Vision for Explainable Automated Algorithm Design}
%
%
\titlerunning{From Performance to Understanding}
\author{Niki van Stein\inst{1}~\orcidlink{0000-0002-0013-7969} \and Anna V. Kononova\inst{1}~\orcidlink{0000-0002-4138-7024} \and Thomas B{\"a}ck\inst{1}~\orcidlink{0000-0001-6768-1478}}

\authorrunning{N. van Stein et al.}

\institute{LIACS, Leiden University, The Netherlands \\
\email{\texttt{\{n.van.stein,a.kononova,t.h.w.baeck\}@liacs.leidenuniv.nl}}}

\maketitle
\begin{abstract}
Automated algorithm design is entering a new phase: Large Language Models can now generate full optimisation (meta)heuristics, explore vast design spaces and adapt through iterative feedback. Yet this rapid progress is largely performance-driven and opaque. Current LLM-based approaches rarely reveal why a generated algorithm works, which components matter or how design choices relate to underlying problem structures. This paper argues that the next breakthrough will come not from more automation, but from coupling automation with understanding from systematic benchmarking. We outline a vision for explainable automated algorithm design, built on three pillars: (i) LLM-driven discovery of algorithmic variants, (ii) explainable benchmarking that attributes performance to components and hyperparameters and (iii) problem-class descriptors that connect algorithm behaviour to landscape structure. Together, these elements form a closed knowledge loop in which discovery, explanation and generalisation reinforce each other. We argue that this integration will shift the field from blind search to interpretable, class-specific algorithm design, accelerating progress while producing reusable scientific insight into when and why optimisation strategies succeed.
\end{abstract}

\begin{figure}[!b]
  \centering
  \includegraphics[width=1.0\textwidth,trim=26mm 27mm 26mm 31mm,clip]{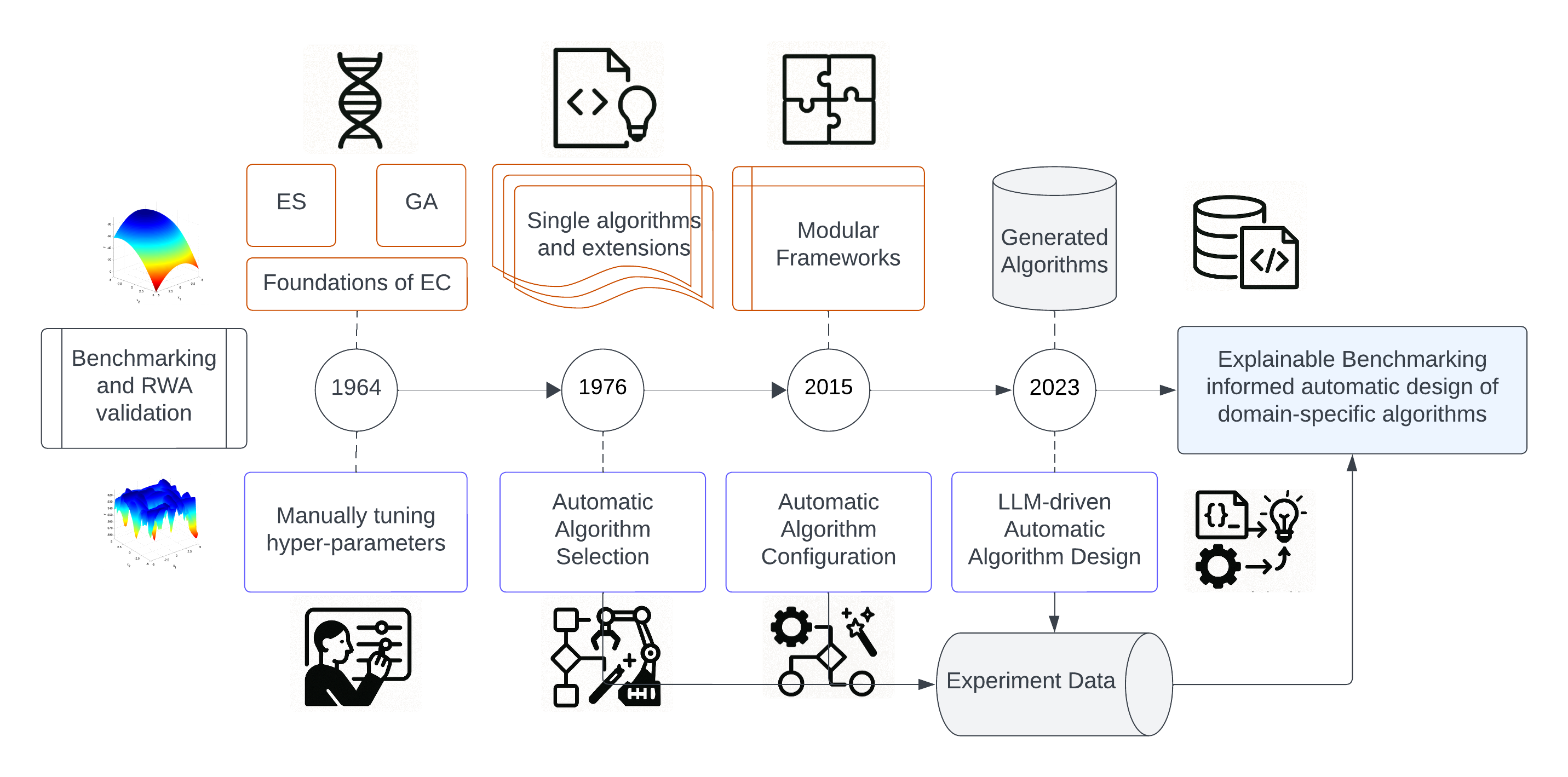}
  \caption{General timeline of algorithm development in Evolutionary Computation\\ -- a vision.}
  \label{fig:vision}
\end{figure}

\section{Introduction}
Evolutionary computation (EC) has steadily progressed toward greater \emph{automation and abstraction}: From hand-crafted heuristics, through hyper-parameter optimisation (HPO), algorithm selection and configuration, modular algorithm spaces and now to fully automated algorithm \emph{design} enabled by large language models (LLMs). This shift expands the design space and reduces human bottlenecks, but it also exposes two pressing needs:  
(i) \textbf{explainable benchmarking}, to understand \emph{why} and \emph{when} algorithmic components and hyperparameters matter, 
(ii) \textbf{problem (and instance) descriptors} that capture structural properties of problem \emph{classes} as well as individual problem \emph{instances}.  
As work in combinatorial optimisation illustrates, even within a single problem class such as the Travelling Salesperson Problem, instances can vary dramatically in difficulty and structure and meaningful algorithm design must account for these differences.

We argue that combining LLM-driven discovery with explainable benchmarking and principled landscape descriptors offers a path toward \emph{class- and instance-specific automated algorithm discovery}. Such an integration promises not only more effective solvers, but also a deeper scientific understanding of how algorithmic components interact with problem structure, ultimately enabling scalable, interpretable and use-case-specific algorithm design.


\section{Past and Current Trends in EC}
\subsection{A Brief History of Algorithm Design in EC}

Evolutionary Computation dates back to the 1960s (see Figure~\ref{fig:vision}), when \emph{Genetic Algorithms} (GA)~\cite{Holland75}, \emph{Evolutionary Programming} (EP)~\cite{Fogel1965} and \emph{Evolutionary Strategies} (ES)~\cite{Rec73,schwefel1977numerische} were proposed in the US (GA, EP), as well as in Germany (ES). 
Ever since their invention, their main focus has been on finding near-optimal solutions to non-linear black-box optimisation problems $\min f: {\cal D} \rightarrow \mathbb{R}^k$, defined over some domain ${\cal D}$, initially for the single-objective case $k=1$ (e.g., see \cite{BaeckS93,DBLP:series/ncs/EibenS03,DBLP:books/daglib/0082827}), later on also for multi-objective problems ($k>1$) (e.g., \cite{Deb09}).

In the late 1980s, automated parameter tuning by a meta-level GA has been introduced into the field \cite{Gre86}, later on followed by combined operator selection and parameter tuning \cite{Bae94}, as a precurser to what is being called \emph{hyper-parameter optimisation} and \emph{algorithm configuration} today (e.g., see \cite{DBLP:journals/air/BaratchiWLRHBO24} for an overview). 

An alternative method, the continuous internal adaptation of certain hyperparameters through a process was called \emph{self-adaptation} \cite{DBLP:journals/evi/Kramer10,schwefel1981numerical}, has been introduced in ES from the beginning, serving as an inspiration over decades towards the \emph{covariance matrix adaptation} (CMA-)ES \cite{hansen2016cmatut,hansen2015evolution} and variants thereof \cite{BFK13}.

From the early 1990s, the development of new variants of such algorithms also accelerated significantly due to the integration of the three mainstream methods into the overarching field of \emph{Evolutionary Computation} (EC), enabling hybridisation and cross-fertilisation among those previously separate developments (see \cite{DBLP:books/daglib/0092410} or the later version \cite{back2023evolutionary} for some examples). Hybridisation took highly sophisticated forms, most notably in \textit{adaptive memetic algorithms}~\cite{Moscato1989_report,Krasnogor2002_thesis}, which coordinate multiple local search methods tailored to different problem landscapes or optimisation stages, through complex adaptive rules~\cite{Ong2006}. These mechanisms allow algorithms to adjust their behaviour to diverse landscape characteristics, thereby extending their ability to address increasingly demanding optimisation tasks, with a relatively limited set of coordinated algorithmic components.

Ever more sophisticated algorithms and variations thereof have been proposed by experts in the field, up to a point where even in a subfield such as ES a plethora of algorithmic variants exist \cite{BFK13}. 
Beyond the field of EC, which now has a firm theoretical foundation based on decades of convergence theory and runtime  \cite{DBLP:series/ncs/2020DN,DoerrN21}, a huge number of nature-inspired metaheuristics has been proposed.
However, these are often insufficiently benchmarked (ignoring state-of-the-art algorithms), insufficiently formalised (using imprecise algorithmic and no mathematical formalism) or are minor variations of existing methods - often not even able to compete with random search (see e.g.~\cite{DBLP:journals/itor/CamachoVillalonDS23,camacho2020grey,CamachoVillalonD22,sorensen2015metaheuristics} for a discussion and \cite{DBLP:conf/gecco/VermettenD0KB24} for a comparison against random search).

As a consequence of the manual, unstructured and incremental process of algorithm development, some researchers started around 2015 to 
propose modularized algorithm design frameworks in which the combinatorics of algorithmic modules (e.g., many existing variants of mutation, recombination, selection and parameter adaptation operators) is implemented in a way that allows complete enumeration or algorithmic search to be applied to these algorithm configuration design spaces. 
This approach resulted in the insight that one could systematically find significantly improved variants of algorithms in fields such as differential evolution \cite{DBLP:conf/gecco/VermettenCKB23}, ES \cite{rijn2016evolving}, particle swarm optimisation \cite{boks2020modular,CVDS22}, differential evolution \cite{DBLP:conf/gecco/VermettenCKB23}, multi-objective algorithms \cite{DBLP:journals/ec/BezerraLS20} and heuristics \cite{DBLP:journals/eor/MartinSantamariaLSC24},
to name the most prominent instances of such modular frameworks. 

If one considers the design of optimisation algorithms to be an optimisation process by itself, the development of Large Language Models (LLM) that can generate program code enables a natural step beyond modular frameworks by allowing to generate optimisation algorithm code from scratch.
Since the LLM has been trained on existing code, it re-uses existing algorithms and can combine and vary program snippets (similar to modules in a modular framework) thereof in novel ways.
This new field of \textit{LLM-driven automated algorithm design in the loop} (a - typically evolutionary - improvement loop is required) started in 2023 with a number of  approaches proposed in parallel (e.g., 
FunSearch \cite{DBLP:journals/nature/RomeraParedesBNBKDREWFKF24}; Large Language Model Evolutionary Algorithm, LLaMEA \cite{DBLP:journals/corr/abs-2405-20132,DBLP:journals/tec/SteinB25}; Evolution of Heuristics, EoH \cite{DBLP:conf/icml/0044TY0LWL024,DBLP:conf/aaai/Yao00L0025}).

It should be noted that a necessary ingredient for any proper algorithm development - whether manual or automated - is the availability of a sophisticated, unbiased, statistically sound and automated \textit{benchmarking tool} ${\cal B}$, as only this allows the quality of an optimisation algorithm $\cal A$ to be evaluated by a single scalar performance measure ${\cal B}(\cal A, \cal F) \in \mathbb{R} \rightarrow \max$
(which is often in practice normalised to be in the interval $[0,1]$), aggregated on a set $\cal F$ of test function instances.
The most relevant benchmarking platforms are COCO \cite{hansen2021coco}, Nevergrad \cite{bennet2021nevergrad} and IOHprofiler \cite{de2021iohexperimenter,wang2022iohanalyzer}.
Due to its relevance, benchmarking is discussed in more detail in the next section.

\subsection{Evolution of Benchmarking Practices} \label{subsec:benchmarking}
Benchmarking plays a central role in the evaluation and development of optimisation algorithms. In its modern form, benchmarking is understood as a systematic assessment of algorithmic performance on a carefully selected set of problem instances, using clearly defined metrics to compare efficiency, accuracy and robustness under varying conditions~\cite{vanStein2025_xplainer}. Historically, benchmarking practices have evolved through several distinct stages.

\paragraph{Demonstration of applicability of the algorithm} Early benchmarking focused on demonstrating that an algorithm worked on a particular class of problems, typically by plotting average fitness over time across several independent runs. These studies rarely included competing methods and often relied on qualitative, ‘verdict-like’ statements about applicability. This later evolved into comparisons of a small number of algorithms on emerging benchmark suites, although conclusions still tended to rely heavily on visual inspection rather than rigorous analysis.

\paragraph{Extensive convergence plots} As encouraged by emerging tools such as COCO~\cite{hansen2009real} and workshops such as the ACM GECCO Black Box Optimization Benchmarking series, a more extensive benchmarking strategy relies on visual analysis of standardised convergence plots for all functions in a suite. While this offers broader insight, interpreting many plots requires expertise and is sometimes complemented—although not always convincingly—by statistical tests or by summary tables that limit interpretability.

\paragraph{Performance per function group} Benchmarking practices increasingly consider performance per function group, using high-level properties (e.g. multimodality, global structure, separability) or landscape features obtained via Exploratory Landscape Analysis \cite{mersmann2011exploratory,munoz2015alg} to understand when algorithms succeed or fail.

\paragraph{Domain-specfic benchmarking} Domain-specific benchmarks highlight that standard suites may not reflect real applications, risking overfitting~\cite{blomDVMNNO20}. This has motivated specialised algorithms for particular domains~\cite{kohira2018,Tanabe2020}.

\paragraph{Modern benchmarking} While benchmarking has become somewhat more rigorous, modern practices~\cite{bartz2020benchmarking} still face several limitations: (i) ablation studies are often absent, (ii) performance aggregation typically assumes uniform problem distributions~\cite{hansen2021coco}, while metrics such as absolute runtime distributions or performance profiles may obscure important differences in scalability. Explainability also remains limited, although initial progress has been made through analyses of algorithm complementarity and efforts to relate performance to problem features, for example via instance space analysis~\cite{Smith-Miles2014} or more structured methods for extracting semantic relations~\cite{Kostovska2022}.

\paragraph{Explainable Benchmarking} Recently, a new framework for experimentation and analysis of black-box iterative optimisation heuristics called IOHxplainer has been proposed based on the following~\cite{vanStein2025_xplainer}:

\begin{description}
    \item[Explainable Benchmarking] formalises the idea of linking an algorithm’s configuration and a benchmark problem to the performance it achieves, enriched with explainability metrics. It considers a family of algorithms, a chosen subset of configurations, a benchmarking suite and an experimental setup and seeks to describe performance as a function of both algorithm parameters and problem characteristics. \item[Explainable Prediction] constitutes a more ambitious goal, where such mappings could be used to anticipate performance on unseen algorithms, unseen problems or untested experimental setups, although achieving this fully remains beyond current computing and methodological capabilities.
\end{description}

\subsection{Automated Algorithm Selection, Configuration and Design}
Automated Algorithm Selection (AAS) and Automated Algorithm Configuration (AAC) form the foundational layers of modern algorithmic automation.
The AAS paradigm was introduced by Rice in 1976 \cite{rice1976algorithm} as the problem of mapping instance features to algorithm performance to select the most appropriate solver for each task.
Over the past decades, this framework evolved from hand-crafted meta-models to machine-learning–based selectors that learn performance mappings directly from data \cite{kerschke2019automated}.
Within evolutionary computation, early AAS work incorporated Exploratory Landscape Analysis features \cite{mersmann2011exploratory} to characterise problem instances and enable per-instance algorithm selection \cite{kerschke2019automatedBlack}.
New exploratory landscape analysis approaches such as Deep-ELA \cite{seiler2025deep} highlight progress towards large-scale and explainable selection models that integrate deep representations of problem structures.

In parallel, AAC seeks to automatically identify the best configuration of hyperparameters for a given algorithm and problem class.
Frameworks such as ParamILS \cite{hutter2009paramils}, SMAC \cite{lindauer2022smac3} and irace \cite{lopez2016irace} established the statistical foundations of configuration as a black-box optimisation problem.
Evolutionary and surrogate-based extensions (e.g., SPOT \cite{bartz2007experimental}, Hyperband \cite{li2018hyperband}) further improved scalability and efficiency.
Reviews \cite{hoos2012automated} emphasise that AAC bridges automated tuning and meta-learning by discovering configuration policies transferable across problem classes.

Together, AAS and AAC underpin the transition towards fully Automated Algorithm Design (AAD).
They formalise the learning of algorithm–problem relations and provide the statistical infrastructure upon which modern generative approaches, such as LLM-driven algorithm design, can build.
The evolution of AAD can be traced back to the broader development of evolutionary computation and the increasing automation of algorithmic discovery (see Figure~\ref{fig:vision}). 
Early explorations into self-modifying systems emerged in the 1970s with \textit{Eurisko} \cite{lenat1983eurisko}, one of the first attempts to enable computers to invent their own heuristics.
This idea of algorithmic self-improvement was later grounded in evolutionary principles through the work on genetic programming by Koza in the 1990s \cite{o1994genetic}, 
which demonstrated how evolutionary operators could evolve complete programs and algorithmic structures.

In the early 2000, the field of \textit{hyper-heuristics} formalised the idea of ``heuristics to choose heuristics'' \cite{burke2003hyper}, 
marking a conceptual shift from parameter optimisation to the automated generation and selection of heuristic strategies. 
Throughout the 2010s, research began to focus on the composition and assembly of heuristics, for example through self-assembling design processes \cite{terrazas2010towards}, 
and surveys started consolidating the emerging methods under the umbrella of automated heuristic design \cite{ochoa2011automated}. 

In the following decade, general frameworks capable of producing metaheuristic algorithms with diverse internal structures appeared, such as AutoOpt \cite{zhao2025autoopt}, 
which pushed the boundary beyond configuration towards true algorithm generation. 
More recent surveys \cite{stutzle2018automated,zhao2023automated} positioned AAD as a distinct research field, bridging evolutionary computation, machine learning and software synthesis.

Today, the frontier of AAD integrates large language models as creative agents capable of generating novel algorithmic components or even entire algorithmic frameworks \cite{DBLP:conf/icml/0044TY0LWL024,DBLP:journals/corr/abs-2506-13131,DBLP:journals/nature/RomeraParedesBNBKDREWFKF24,DBLP:journals/tec/SteinB25,ye2024reevo}.
This trajectory, from handcrafted, hand-tuned heuristics to LLM-driven algorithm synthesis illustrates a clear progression toward automation and abstraction in the design of intelligent search processes.

\subsection{Motivation}
The evolution of algorithm design has followed a cyclical pattern of invention, refinement and automation. We have entered a \textit{new iteration} of this cycle, but now on a higher level of abstraction: rather than designing individual algorithms, we design \emph{methods that design algorithms}. AAD systems, now often driven by LLMs, promise rapid exploration of the open-ended algorithmic search space and continuous improvement through feedback. However, as with every previous wave of automation, new challenges emerge.

Without \emph{explanation} and \emph{understanding}, automation risks turning into blind exploration (e.g. random search). While recent AAD frameworks demonstrate impressive generative capabilities, they rarely provide insight into \emph{why} a generated algorithm performs well, \emph{which} of its components are responsible, or \emph{how} it relates to the underlying problem characteristics. This limits both reproducibility and scientific understanding (discovery). The field therefore needs systematic methods to attribute performance to algorithmic design choices and hyperparameters and to link these attributions to structural properties of the problems being solved.

\emph{Explainable benchmarking} and problem-class characterisation can close this gap. By embedding attribution mechanisms and problem descriptors into the AAD pipeline, we can move from empirical discovery toward \emph{interpretable discovery}. Such integration would not only accelerate progress through data-driven feedback loops but also generate reusable design knowledge, bridging the current divide between automated synthesis and human understanding.

\textbf{Our motivation.} This paper argues for combining LLM-driven discovery with explainable benchmarking and problem-class descriptors to enable class-specific, interpretable algorithm design. We advocate a transition from opaque performance improvement to transparent, explainable progress, where each algorithmic innovation contributes to a broader understanding of how structure, performance and design interrelate.

\section{The Case for Class-Specific Discovery}
As prescribed by the so-called ``No Free Lunch'' theorems (NFLTs)~\cite{NFLT,Rowe2009}, \textit{no} universally superior optimisation algorithm exists. The idea that one might identify the best overall algorithm across all possible problems is therefore devoid of meaning. When \textit{no} assumptions are made about the structure of the problem class the algorithm is to be applied to, no optimisation algorithm can outperform any other in a universal sense. In such a setting, no algorithm can obtain a performance advantage on its own. Any advantage becomes possible only when the problem class is restricted, as this restriction implicitly introduces structure that some algorithms can exploit. By restricting attention to classes of functions with specified regularities or constraints, we move beyond the overly general setting assumed by the NFLTs. These more narrowly defined problem classes may theoretically permit free lunches. A clear example is continuous optimisation, where NFL does not apply because the assumptions required by the theorem cannot be satisfied by continuous function classes~\cite{Auger2010} and within such settings, some algorithms can indeed be shown to perform better on average.

Introducing additional structure or specialisation creates further opportunities for free lunches. For example, free-lunch phenomena emerge when one considers averages over structured multi-objective problem classes~\cite{Corne2003_no,Corne2003_some}. Similar effects arise in co-evolutionary contexts, where interacting populations and coupled objective landscapes induce statistical regularities that can be exploited~\cite{Wolpert2005_coevolutionary}. These examples reinforce the view that meaningful performance differences become visible only when the problem class departs from the fully unrestricted case.

While continuity is sufficient to break the assumptions of the NFLTs, this insight alone has \textit{limited} practical value. The class of continuous functions remains so broad that it offers little guidance for understanding or predicting algorithmic performance. It is therefore necessary to consider narrower and more structured problem classes that better reflect the characteristics of real optimisation tasks. Only within such refined settings can we explain observed performance differences and formulate principled recommendations for algorithm design.

From a class-specific perspective, the goal is to use problem descriptors to identify which algorithmic elements and settings are most suitable for a given problem class with structure. However, imposing structure in practice is not straightforward. Exploratory Landscape Analysis~\cite{mersmann2010benchmarking,mersmann2011exploratory} provides possible principled basis for defining such descriptors. With these in place, discovery can target problem classes~\cite{Long2022_gecco,Thomaser2022_one-shot,deWinter2024}, not only individual problem instances and, thus moving in the direction of reusable design knowledge and selectors~\cite{Long2025_surrogate,Thomaser2023_transfer}.



\section{Integrating Discovery and Explanation: A Vision}
Automated algorithm design is moving from blind exploration to structured discovery. We envision a \emph{closed knowledge loop}:
\begin{enumerate}
  \item \textbf{Discover:} Use LLM-driven search (e.g., LLaMEA\cite{DBLP:journals/tec/SteinB25}) to propose and refine algorithms over real-world inspired benchmark suites.
  \item \textbf{Explain:} Run explainable benchmarking techniques to attribute performance to components and settings across different problem classes and instances.
  \item \textbf{Describe:} Learn problem descriptors that align with observed attributions and cluster functions into classes.
  \item \textbf{Generalise:} Induce class-specific design rules and selectors; feed these rules back into prompts, mutation policies and priors for the next discovery cycle.
\end{enumerate}

In this loop, large language models act as \textit{creative engines}. They generate, mutate and refine optimisation algorithms from natural language prompts or feedback. Frameworks such as LLaMEA~\cite{DBLP:journals/tec/SteinB25} and EoH~\cite{DBLP:conf/icml/0044TY0LWL024} operationalise this idea by combining LLM-driven code generation with evolutionary selection.  Candidate algorithms are rigorously evaluated on benchmark suites, their performance aggregated through measures such as area over the convergence curve or gap to the known best and improved iteratively based on the best-performing designs. To focus LLM capacity on structural innovation rather than numeric tuning, hybrid setups such as LLaMEA-HPO~\cite{van2024loop} delegate hyper-parameter optimisation to specialised tools like SMAC \cite{lindauer2022smac3}, improving both efficiency and scalability.

A future accelerator of this loop could be a tighter integration of \emph{explainable benchmarking}~\cite{back2023evolutionary,vanStein2025_xplainer}. Traditional benchmarking quantifies performance; explainable benchmarking interprets it. It attributes performance differences to algorithmic components, hyperparameters and their interactions across problem instances and classes. Frameworks such as IOHxplainer~\cite{vanStein2025_xplainer} build surrogate models over large configuration–performance datasets and apply explainable AI techniques or sensitivity analyses to reveal which parts of an algorithm contribute most to success. This process transforms experimental data into \textit{actionable} insight, identifying for example, when self-adaptation or recombination operators matter most and why. These insights can be used to steer the LLM-driven search by for example modifying mutation and selection procedures.

Good \textit{problem descriptors} can then close the loop by providing structure on the problem side. ELA~\cite{mersmann2011exploratory} and its recent deep extensions (e.g., Deep-ELA~\cite{seiler2025deep}) can capture landscape features that cluster functions into interpretable problem classes. Linking these descriptors with algorithmic attributions allows us to discover not just \emph{what} works, but \emph{where} and \emph{why}. 

Together, these elements form an iterative cycle: LLMs discover new algorithms, explainable benchmarking analyses them and problem descriptors generalise the findings into reusable design knowledge. The insights gained can then be embedded back into LLM prompts, mutation operators or priors, guiding the next generation of discoveries. This vision shifts automated algorithm design from purely performance based search toward an interpretable, data-driven \textit{science of algorithmic behaviour}. Each iteration not only produces better solvers but also deepens our understanding of the principles behind them.





\section{Implications and Research Agenda}

The vision outlined above has concrete methodological and scientific consequences for the field.  
If automated algorithm design is to become \emph{class-specific, explainable and reusable}, several research directions must be prioritised.

\begin{description}
    \item[Landscape features and problem descriptors] A central requirement is the development of richer, more discriminative problem descriptors.  
    Exploratory Landscape Analysis has shown that structural regularities can be extracted from black-box problems, but current descriptors remain limited in scale, sensitivity and real-world applicability.  
    Progress here should be driven by realistic or real-world–inspired problem sets, enabling descriptors that generalise beyond synthetic landscapes.  
    Improved descriptors will directly strengthen AAS models and enable transfer learning within AAD, allowing design knowledge to move across related problem classes.

    \item[Advances in AAS, AAD and retrieval-augmented discovery]
    Better problem representations will push developments in automated algorithm selection and configuration.  
    Machine-learning models for AAS can be improved by integrating learned descriptors, while AAD pipelines can benefit from retrieval-augmented generation mechanisms that condition LLMs on prior discoveries, structural rules or examples drawn from known problem classes.  
    This requires systematic work on how to embed algorithm–problem relations into the discovery loop.
    
    \item[Attribution and explainable benchmarking]
    To move beyond blind exploration, the community needs more robust methods for attributing performance to algorithmic components, hyperparameters and their interactions.  
    Explainable benchmarking should become a standard layer in AAD pipelines, providing principled sensitivity analyses, component-level importance scores and interaction effects across problem classes.  
    Such procedures not only guide discovery but also generate reusable scientific insight.
    
    \item[Encoding design knowledge into prompts]
    Another research direction concerns how to inject algorithmic knowledge into LLM prompts or mutation operators.  
    Encoding rules, motifs, constraints or high-level design principles into prompts can shape the search space and reduce wasted exploration.  
    Equally important is distinguishing between structural innovation (algorithmic modules, operator choices, control mechanisms) and numeric tuning; delegating the latter to specialised HPO tools remains essential for scala\-bi\-lity.
    
    \item[Standardisation, benchmarking and tooling]
    The community must invest in shared protocols and tooling.  
    Standard evaluation budgets, anytime performance metrics, aggregation rules and reporting templates will make results comparable and reproducible.  
    Tooling such as IOH~\cite{wang2022iohanalyzer}, BLADE~\cite{BLADE} and LLM4AD~\cite{liu2024llm4ad} already provide strong foundations, but widespread adoption requires standard mechanisms for storing, publishing and sharing large benchmarking datasets and experiment metadata and results. Moreover, benchmarking suites need to be developed specifically for AAS and AAD.
    
    \item[From empirical insight to theory]
    Finally, attribution patterns must be map\-ped back to known algorithmic mechanisms.  
    This creates an opportunity to test falsifiable hypotheses~\cite{Popper1959_logic,Eiben2002_critical} about when and why specific components matter.  
    If done systematically, explainable benchmarking can feed the growth of new theoretical results grounded in empirical regularities, closing the loop between data-driven discovery and proof-driven understanding.
\end{description}

Overall, this agenda shifts the field from purely empirical improvement toward a \textit{principled science} of algorithmic behaviour, where automated discovery, structured explanation and theoretical insight reinforce one another.

\section{Conclusion}


The field is poised to move from automated tuning to \emph{explainable, problem-class-specific} algorithm discovery. LLM-driven design provides exploratory power, explainable benchmarking provides attribution of performance to algorithm components and hyperparameters and finally problem descriptors provide the semantic glue between problems and components. Together, they promise a data-to-design pipeline that learns \emph{which} pieces matter, \emph{why} and \emph{for which} classes, accelerating progress while keeping it interpretable. In doing so, this agenda moves the field beyond purely empirical improvement toward a more \emph{principled science} of algorithmic behaviour, where automated discovery is guided by structural understanding and testable explanations rather than trial and error.
The future of this field is a hybridisation between evolutionary methods, artificial intelligence and real-world inspired benchmarking practices.

\subsubsection*{Disclosure of Interests.}
The authors have no competing interests to declare that are relevant to the content of this article.

\bibliographystyle{splncs04}
\bibliography{references}

@article{kerschke2019automatedBlack,
  title={Automated algorithm selection on continuous black-box problems by combining exploratory landscape analysis and machine learning},
  author={Kerschke, Pascal and Trautmann, Heike},
  journal={Evolutionary computation},
  volume={27},
  number={1},
  pages={99--127},
  year={2019},
  publisher={MIT Press One Rogers Street, Cambridge, MA 02142-1209, USA journals-info~…}
}

@article{lenat1983eurisko,
  title={EURISKO: a program that learns new heuristics and domain concepts: the nature of heuristics III: program design and results},
  author={Lenat, Douglas B},
  journal={Artificial intelligence},
  volume={21},
  number={1-2},
  pages={61--98},
  year={1983},
  publisher={Elsevier}
}

@article{liu2024llm4ad,
  title={Llm4ad: A platform for algorithm design with large language model},
  author={Liu, Fei and Zhang, Rui and Xie, Zhuoliang and Sun, Rui and Li, Kai and Lin, Xi and Wang, Zhenkun and Lu, Zhichao and Zhang, Qingfu},
  journal={arXiv preprint arXiv:2412.17287},
  year={2024}
}

@inproceedings{BLADE,
author = {van Stein, Niki and V. Kononova, Anna and Yin, Haoran and B\"{a}ck, Thomas},
title = {{BLADE}: Benchmark suite for LLM-driven Automated Design and Evolution of iterative optimisation heuristics},
year = {2025},
isbn = {9798400714641},
publisher = {Association for Computing Machinery},
address = {New York, NY, USA},
url = {https://doi.org/10.1145/3712255.3734347},
doi = {10.1145/3712255.3734347},
booktitle = {Proceedings of the Genetic and Evolutionary Computation Conference Companion},
pages = {2336–2344},
numpages = {9},
location = {NH Malaga Hotel, Malaga, Spain},
series = {GECCO '25 Companion}
}

@article{van2024loop,
  title={In-the-loop hyper-parameter optimization for llm-based automated design of heuristics},
  author={van Stein, Niki and Vermetten, Diederick and B{\"a}ck, Thomas},
  journal={ACM Transactions on Evolutionary Learning},
  year={2024},
  publisher={ACM New York, NY}
}

@article{ye2024reevo,
  title={Reevo: Large language models as hyper-heuristics with reflective evolution},
  author={Ye, Haoran and Wang, Jiarui and Cao, Zhiguang and Berto, Federico and Hua, Chuanbo and Kim, Haeyeon and Park, Jinkyoo and Song, Guojie},
  journal={Advances in neural information processing systems},
  volume={37},
  pages={43571--43608},
  year={2024}
}

@article{zhao2023automated,
  title={Automated design of metaheuristic algorithms: A survey},
  author={Zhao, Qi and Duan, Qiqi and Yan, Bai and Cheng, Shi and Shi, Yuhui},
  journal={arXiv preprint arXiv:2303.06532},
  year={2023}
}

@incollection{stutzle2018automated,
  title={Automated design of metaheuristic algorithms},
  author={St{\"u}tzle, Thomas and L{\'o}pez-Ib{\'a}{\~n}ez, Manuel},
  booktitle={Handbook of metaheuristics},
  pages={541--579},
  year={2018},
  publisher={Springer}
}

@article{zhao2025autoopt,
  author={Zhao, Qi and Yan, Bai and Hu, Taiwei and Chen, Xianglong and Yang, Jian and Cheng, Shi and Shi, Yuhui},
  journal={IEEE Transactions on Emerging Topics in Computational Intelligence}, 
  title={AutoOpt: A General Framework for Automatically Designing Metaheuristic Optimization Algorithms With Diverse Structures}, 
  year={2025},
  volume={9},
  number={5},
  pages={3690-3703},
  doi={10.1109/TETCI.2025.3561629}
}

@inproceedings{ochoa2011automated,
  title={Automated heuristic design},
  author={Ochoa, Gabriela and Hyde, Matthew V and Burke, Edmund K},
  booktitle={Proceedings of the 13th annual conference companion on Genetic and evolutionary computation},
  pages={1321--1342},
  year={2011}
}

@article{terrazas2010towards,
  title={Towards the design of heuristics by means of self-assembly},
  author={Terrazas, German and Landa-Silva, Dario and Krasnogor, Natalio},
  journal={arXiv preprint arXiv:1006.1681},
  year={2010}
}

@incollection{burke2003hyper,
  title={Hyper-heuristics: An emerging direction in modern search technology},
  author={Burke, Edmund and Kendall, Graham and Newall, Jim and Hart, Emma and Ross, Peter and Schulenburg, Sonia},
  booktitle={Handbook of metaheuristics},
  pages={457--474},
  year={2003},
  publisher={Springer}
}

@article{o1994genetic,
  title={Genetic programming II: automatic discovery of reusable programs},
  author={O'Reilly, Una-May},
  journal={Artificial Life},
  volume={1},
  number={4},
  pages={439--441},
  year={1994},
  publisher={MIT Press}
}

@incollection{hoos2012automated,
  title={Automated algorithm configuration and parameter tuning},
  author={Hoos, Holger H},
  booktitle={Autonomous search},
  pages={37--71},
  year={2012},
  publisher={Springer}
}

@article{li2018hyperband,
  title={Hyperband: A novel bandit-based approach to hyperparameter optimization},
  author={Li, Lisha and Jamieson, Kevin and DeSalvo, Giulia and Rostamizadeh, Afshin and Talwalkar, Ameet},
  journal={Journal of Machine Learning Research},
  volume={18},
  number={185},
  pages={1--52},
  year={2018}
}

@inproceedings{bartz2007experimental,
  title={Experimental research in evolutionary computation},
  author={Bartz-Beielstein, Thomas and Preuss, Mike},
  booktitle={Proceedings of the 9th annual conference companion on genetic and evolutionary computation},
  pages={3001--3020},
  year={2007}
}

@article{hutter2009paramils,
  title={ParamILS: an automatic algorithm configuration framework},
  author={Hutter, Frank and Hoos, Holger H and Leyton-Brown, Kevin and St{\"u}tzle, Thomas},
  journal={Journal of artificial intelligence research},
  volume={36},
  pages={267--306},
  year={2009}
}

@article{seiler2025deep,
  title={Deep-ela: Deep exploratory landscape analysis with self-supervised pretrained transformers for single-and multi-objective continuous optimization problems},
  author={Seiler, Moritz Vinzent and Kerschke, Pascal and Trautmann, Heike},
  journal={Evolutionary Computation},
  pages={1--27},
  year={2025},
  publisher={MIT Press 255 Main Street, 9th Floor, Cambridge, Massachusetts 02142, USA~…}
}

@book{BFK13,
  title={Contemporary Evolution Strategies},
  author={B{\"a}ck, T. and Foussette, C. and Krause, P.},
  year={2013},
  publisher={Springer}
}

@book{Rec73,
  title={Evolutionsstrategie: {O}ptimierung technischer {S}ystem nach {P}rinzipien der biologischen {E}volution},
  author={Rechenberg, I.},
  year={1973},
  publisher={frommann-holzboog, Stuttgart}
}

@book{Deb09,
  title={Multi-objective {O}ptimization using {E}volutionary {A}lgorithms},
  author={Deb, Kalyanmoy},
  year={2009},
  publisher={Wiley, NY}
}

@article{BaeckS93,
    author = {B{\"a}ck, Thomas and Schwefel, Hans-Paul},
    title = {An Overview of Evolutionary Algorithms for Parameter Optimization},
    journal = {Evolutionary Computation},
    volume = {1},
    number = {1},
    pages = {1-23},
    year = {1993},
    month = {03},
    issn = {1063-6560}
}

@article{CamachoVillalonD22,
    title = {An analysis of why cuckoo search does not bring any novel ideas to optimization},
    journal = {Computers $\&$ Operations Research},
    volume = {142},
    pages = {105747},
    year = {2022},
    issn = {0305-0548},
    doi = {https://doi.org/10.1016/j.cor.2022.105747},
    url = {https://www.sciencedirect.com/science/article/pii/S0305054822000442},
    author = {Christian L. Camacho-Villalón and Marco Dorigo and Thomas St{\"u}tzle}
}

@book{Holland75, 
author = {Holland, John H.}, 
title = {Adaptation in Natural and Artificial Systems: An Introductory Analysis with Applications to Biology, Control and Artificial Intelligence}, 
year = {1992, 1st edition: 1975}, 
isbn = {0262082136}, 
publisher = {MIT Press}, 
address = {Cambridge, MA, USA}
}

@article{bartz2020benchmarking,
  author    = {Thomas Bartz{-}Beielstein and
               Carola Doerr and
               Jakob Bossek and
               Sowmya Chandrasekaran and
               Tome Eftimov and
               Andreas Fischbach and
               Pascal Kerschke and
               Manuel L{\'{o}}pez{-}Ib{\'{a}}{\~{n}}ez and
               Katherine M. Malan and
               Jason H. Moore and
               Boris Naujoks and
               Patryk Orzechowski and
               Vanessa Volz and
               Markus Wagner and
               Thomas Weise},
  title     = {Benchmarking in Optimization: Best Practice and Open Issues},
  journal   = {CoRR},
  volume    = {abs/2007.03488},
  year      = {2020},
  url       = {https://arxiv.org/abs/2007.03488},
  eprinttype = {arXiv},
  eprint    = {2007.03488},
  bibsource = {dblp computer science bibliography, https://dblp.org}
}

@incollection{hansen2015evolution,
  title={Evolution strategies},
  author={Hansen, Nikolaus and Arnold, Dirk V and Auger, Anne},
  booktitle={Springer handbook of computational intelligence},
  pages={871--898},
  year={2015},
  publisher={Springer}
}

@article{blomDVMNNO20,
  title={Identifying properties of real-world optimisation problems through a questionnaire},
  author={van der Blom, Koen and Deist, Timo M and Volz, Vanessa and Marchi, Mariapia and Nojima, Yusuke and Naujoks, Boris and Oyama, Akira and Tu{\v{s}}ar, Tea},
  journal={arXiv preprint arXiv:2011.05547},
  year={2020}
}

@article{hansen2021coco,
  title={{COCO}: A platform for comparing continuous optimizers in a black-box setting},
  author={Hansen, Nikolaus and Auger, Anne and Ros, Raymond and Mersmann, Olaf and Tu{\v{s}}ar, Tea and Brockhoff, Dimo},
  journal={Optimization Methods and Software},
  volume={36},
  number={1},
  pages={114--144},
  year={2021},
  publisher={Taylor \& Francis}
}

@article{wang2022iohanalyzer,
  title={{IOH}analyzer: Detailed performance analyses for iterative optimization heuristics},
  author={Wang, Hao and Vermetten, Diederick and Ye, Furong and Doerr, Carola and B{\"a}ck, Thomas},
  journal={ACM Transactions on Evolutionary Learning and Optimization},
  volume={2},
  number={1},
  pages={1--29},
  year={2022},
  publisher={ACM New York, NY}
}

@article{bennet2021nevergrad,
  title={Nevergrad: black-box optimization platform},
  author={Bennet, Pauline and Doerr, Carola and Moreau, Antoine and Rapin, Jeremy and Teytaud, Fabien and Teytaud, Olivier},
  journal={ACM SIGEVOlution},
  volume={14},
  number={1},
  pages={8--15},
  year={2021},
  publisher={ACM New York, NY, USA}
}

@article{de2021iohexperimenter,
  title={{IOH}experimenter: Benchmarking Platform for Iterative Optimization Heuristics},
  author={de Nobel, Jacob and Ye, Furong and Vermetten, Diederick and Wang, Hao and Doerr, Carola and B{\"a}ck, Thomas},
  journal={arXiv preprint arXiv:2111.04077},
  year={2021}
}

@inproceedings{mersmann2010benchmarking,
  title={Benchmarking evolutionary algorithms: Towards exploratory landscape analysis},
  author={Mersmann, Olaf and Preuss, Mike and Trautmann, Heike},
  booktitle={International Conference on Parallel Problem Solving from Nature},
  pages={73--82},
  year={2010},
  organization={Springer}
}

@misc{hansen2009real,
  title={Real-parameter black-box optimization benchmarking 2009: Noiseless functions definitions},
  author={Hansen, Nikolaus and Finck, Steffen and Ros, Raymond and Auger, Anne},
  year={2009},
  school={INRIA}
}

@inproceedings{mersmann2011exploratory,
  title={Exploratory landscape analysis},
  author={Mersmann, Olaf and Bischl, Bernd and Trautmann, Heike and Preuss, Mike and Weihs, Claus and Rudolph, G{\"u}nter},
  booktitle={Proceedings of Genetic and Evolutionary Computation Conference},
  pages={829--836},
  year={2011},
  series={GECCO '11}
}

@article{lopez2016irace,
  title={The irace package: Iterated racing for automatic algorithm configuration},
  author={L{\'o}pez-Ib{\'a}{\~n}ez, Manuel and Dubois-Lacoste, J{\'e}r{\'e}mie and C{\'a}ceres, Leslie P{\'e}rez and Birattari, Mauro and St{\"u}tzle, Thomas},
  journal={Operations Research Perspectives},
  volume={3},
  pages={43--58},
  year={2016},
  publisher={Elsevier}
}

@article{lindauer2022smac3,
  title={{SMAC3:} A Versatile {B}ayesian Optimization Package for Hyperparameter Optimization.},
  author={Lindauer, Marius and Eggensperger, Katharina and Feurer, Matthias and Biedenkapp, Andr{\'e} and Deng, Difan and Benjamins, Carolin and Ruhkopf, Tim and Sass, Ren{\'e} and Hutter, Frank},
  journal={J. Mach. Learn. Res.},
  volume={23},
  pages={54--1},
  year={2022}
}

@incollection{rice1976algorithm,
  title={The algorithm selection problem},
  author={Rice, John R},
  booktitle={Advances in computers},
  volume={15},
  pages={65--118},
  year={1976},
  publisher={Elsevier}
}

@article{sorensen2015metaheuristics,
  title={Metaheuristics—the metaphor exposed},
  author={S{\"o}rensen, Kenneth},
  journal={International Transactions in Operational Research},
  volume={22},
  number={1},
  pages={3--18},
  year={2015},
  publisher={Wiley Online Library}
}

@inproceedings{camacho2020grey,
  title={Grey wolf, firefly and bat algorithms: Three widespread algorithms that do not contain any novelty},
  author={Camacho Villal{\'o}n, Christian Leonardo and St{\"u}tzle, Thomas and Dorigo, Marco},
  booktitle={International conference on swarm intelligence},
  pages={121--133},
  year={2020},
  organization={Springer}
}

@article{kerschke2019automated,
  title={Automated algorithm selection: Survey and perspectives},
  author={Kerschke, Pascal and Hoos, Holger H and Neumann, Frank and Trautmann, Heike},
  journal={Evolutionary Computation},
  volume={27},
  number={1},
  pages={3--45},
  year={2019},
  publisher={mIT Press}
}

@book{schwefel1981numerical,
  title={Numerical optimization of computer models},
  author={Schwefel, Hans-Paul},
  year={1981},
  publisher={John Wiley \& Sons, Inc.}
}

@inproceedings{boks2020modular,
  author    = {Rick Boks and
               Hao Wang and
               Thomas B{\"{a}}ck},
  editor    = {Carlos Artemio Coello Coello},
  title     = {A modular hybridization of particle swarm optimization and differential
               evolution},
  booktitle = {Genetic and Evolutionary Computation Conference, {GECCO} '20, Companion
               Volume, Canc{\'{u}}n, Mexico, July 8-12, 2020},
  pages     = {1418--1425},
  publisher = {{ACM}},
  year      = {2020},
  url       = {https://doi.org/10.1145/3377929.3398123},
  doi       = {10.1145/3377929.3398123},
  bibsource = {dblp computer science bibliography, https://dblp.org}
}

@inproceedings{Bae94,
  author    = {Thomas B{\"{a}}ck},
  editor    = {Yuval Davidor and
               Hans{-}Paul Schwefel and
               Reinhard M{\"{a}}nner},
  title     = {Parallel Optimization of Evolutionary Algorithms},
  booktitle = {Parallel Problem Solving from Nature - {PPSN} III, International Conference
               on Evolutionary Computation. The Third Conference on Parallel Problem
               Solving from Nature, Jerusalem, Israel, October 9-14, 1994, Proceedings},
  series    = {Lecture Notes in Computer Science},
  volume    = {866},
  pages     = {418--427},
  publisher = {Springer},
  year      = {1994},
  url       = {https://doi.org/10.1007/3-540-58484-6\_285},
  doi       = {10.1007/3-540-58484-6\_285},
  bibsource = {dblp computer science bibliography, https://dblp.org}
}

@ARTICLE{Gre86,
  author={Grefenstette, John J.},
  journal={IEEE Transactions on Systems, Man, and Cybernetics}, 
  title={Optimization of Control Parameters for Genetic Algorithms}, 
  year={1986},
  volume={16},
  number={1},
  pages={122-128},
  doi={10.1109/TSMC.1986.289288}}

@ARTICLE{CVDS22,
  author={Camacho-Villalón, Christian L. and Dorigo, Marco and Stützle, Thomas},
  journal={{IEEE Transactions on Evolutionary Computation}}, 
  title={{PSO-X}: {A} Component-Based Framework for the Automatic Design of Particle Swarm Optimization Algorithms}, 
  year={2022},
  volume={26},
  number={3},
  pages={402-416},
  doi={10.1109/TEVC.2021.3102863}}

@article{Fogel1965,
  author={Fogel, L. J. and Owens, A. J. and Walsh, M. J.},
  journal={IEEE Transactions on Human Factors in Electronics}, 
  title={Intelligent decision-making through a simulation of evolution}, 
  year={1965},
  volume={HFE-6},
  number={1},
  pages={13-23},
  doi={10.1109/THFE.1965.6591252}
}

@article{hansen2016cmatut,
  author    = {Nikolaus Hansen},
  title     = {The {CMA} Evolution Strategy: {A} Tutorial},
  journal   = {CoRR},
  volume    = {abs/1604.00772},
  year      = {2016},
  url       = {http://arxiv.org/abs/1604.00772},
  eprinttype = {arXiv},
  eprint    = {1604.00772},
  bibsource = {dblp computer science bibliography, https://dblp.org}
}

@article{DoerrN21,
  author    = {Benjamin Doerr and
               Frank Neumann},
  title     = {A Survey on Recent Progress in the Theory of Evolutionary Algorithms
               for Discrete Optimization},
  journal   = {{ACM} Transactions on Evolutionary Learning and Optimization},
  volume    = {1},
  number    = {4},
  pages     = {16:1--16:43},
  year      = {2021},
  url       = {https://doi.org/10.1145/3472304},
  doi       = {10.1145/3472304},
  bibsource = {dblp computer science bibliography, https://dblp.org}
}

@INPROCEEDINGS{rijn2016evolving,
  author={van Rijn, Sander and Wang, Hao and van Leeuwen, Matthijs and Bäck, Thomas},
  booktitle={2016 IEEE Symposium Series on Computational Intelligence (SSCI)}, 
  title={Evolving the structure of Evolution Strategies}, 
  year={2016},
  volume={},
  number={},
  pages={1-8},
  doi={10.1109/SSCI.2016.7850138}}

@book{schwefel1977numerische,
  title={Numerische optimierung von computer-modellen mittels der evolutionsstrategie.(Teil 1, Kap. 1-5)},
  author={Schwefel, H-P},
  year={1977},
  publisher={Birkh{\"a}user}
}

@article{back2023evolutionary,
  title={Evolutionary Algorithms for Parameter Optimization—Thirty Years Later},
  author={B{\"a}ck, Thomas HW and Kononova, Anna V and van Stein, Bas and Wang, Hao and Antonov, Kirill A and Kalkreuth, Roman T and de Nobel, Jacob and Vermetten, Diederick and de Winter, Roy and Ye, Furong},
  journal={Evolutionary Computation},
  volume={31},
  number={2},
  pages={81--122},
  year={2023},
  publisher={MIT Press}
}

@book{DBLP:books/daglib/0082827,
  author       = {Zbigniew Michalewicz},
  title        = {Genetic Algorithms + Data Structures = Evolution Programs, Third Revised
                  and Extended Edition},
  publisher    = {Springer},
  year         = {1996},
  url          = {https://doi.org/10.1007/978-3-662-03315-9},
  doi          = {10.1007/978-3-662-03315-9},
  isbn         = {978-3-540-60676-5},
  bibsource    = {dblp computer science bibliography, https://dblp.org}
}

@book{DBLP:series/ncs/EibenS03,
  author       = {A. E. Eiben and
                  James E. Smith},
  title        = {Introduction to Evolutionary Computing},
  series       = {Natural Computing Series},
  publisher    = {Springer},
  year         = {2003},
  url          = {https://doi.org/10.1007/978-3-662-05094-1},
  doi          = {10.1007/978-3-662-05094-1},
  isbn         = {978-3-642-07285-7},
  bibsource    = {dblp computer science bibliography, https://dblp.org}
}

@article{DBLP:journals/air/BaratchiWLRHBO24,
  author       = {Mitra Baratchi and
                  Can Wang and
                  Steffen Limmer and
                  Jan N. van Rijn and
                  Holger H. Hoos and
                  Thomas B{\"{a}}ck and
                  Markus Olhofer},
  title        = {Automated machine learning: past, present and future},
  journal      = {Artif. Intell. Rev.},
  volume       = {57},
  number       = {5},
  pages        = {122},
  year         = {2024},
  url          = {https://doi.org/10.1007/s10462-024-10726-1},
  doi          = {10.1007/S10462-024-10726-1},
  bibsource    = {dblp computer science bibliography, https://dblp.org}
}

@article{DBLP:journals/evi/Kramer10,
  author       = {Oliver Kramer},
  title        = {Evolutionary self-adaptation: a survey of operators and strategy parameters},
  journal      = {Evol. Intell.},
  volume       = {3},
  number       = {2},
  pages        = {51--65},
  year         = {2010},
  url          = {https://doi.org/10.1007/s12065-010-0035-y},
  doi          = {10.1007/S12065-010-0035-Y},
  bibsource    = {dblp computer science bibliography, https://dblp.org}
}

@book{DBLP:books/daglib/0092410,
  author       = {Thomas B{\"{a}}ck},
  title        = {Evolutionary algorithms in theory and practice - evolution strategies,
                  evolutionary programming, genetic algorithms},
  publisher    = {Oxford University Press},
  year         = {1996},
  isbn         = {978-0-19-509971-3},
  bibsource    = {dblp computer science bibliography, https://dblp.org}
}

@inproceedings{DBLP:conf/gecco/VermettenD0KB24,
  author       = {Diederick Vermetten and
                  Carola Doerr and
                  Hao Wang and
                  Anna V. Kononova and
                  Thomas B{\"{a}}ck},
  editor       = {Xiaodong Li and
                  Julia Handl},
  title        = {Large-Scale Benchmarking of Metaphor-Based Optimization Heuristics},
  booktitle    = {Proceedings of the Genetic and Evolutionary Computation Conference,
                  {GECCO} 2024, Melbourne, VIC, Australia, July 14-18, 2024},
  publisher    = {{ACM}},
  year         = {2024},
  url          = {https://doi.org/10.1145/3638529.3654122},
  doi          = {10.1145/3638529.3654122},
  bibsource    = {dblp computer science bibliography, https://dblp.org}
}

@inproceedings{DBLP:conf/gecco/VermettenCKB23,
  author       = {Diederick Vermetten and
                  Fabio Caraffini and
                  Anna V. Kononova and
                  Thomas B{\"{a}}ck},
  editor       = {Sara Silva and
                  Lu{\'{\i}}s Paquete},
  title        = {Modular Differential Evolution},
  booktitle    = {Proceedings of the Genetic and Evolutionary Computation Conference,
                  {GECCO} 2023, Lisbon, Portugal, July 15-19, 2023},
  pages        = {864--872},
  publisher    = {{ACM}},
  year         = {2023},
  url          = {https://doi.org/10.1145/3583131.3590417},
  doi          = {10.1145/3583131.3590417},
  bibsource    = {dblp computer science bibliography, https://dblp.org}
}

@article{DBLP:journals/eor/MartinSantamariaLSC24,
  author       = {Ra{\'{u}}l Mart{\'{\i}}n{-}Santamar{\'{\i}}a and
                  Manuel L{\'{o}}pez{-}Ib{\'{a}}{\~{n}}ez and
                  Thomas St{\"{u}}tzle and
                  J. Manuel Colmenar},
  title        = {On the automatic generation of metaheuristic algorithms for combinatorial
                  optimization problems},
  journal      = {Eur. J. Oper. Res.},
  volume       = {318},
  number       = {3},
  pages        = {740--751},
  year         = {2024},
  url          = {https://doi.org/10.1016/j.ejor.2024.06.001},
  doi          = {10.1016/J.EJOR.2024.06.001},
  bibsource    = {dblp computer science bibliography, https://dblp.org}
}

@article{DBLP:journals/ec/BezerraLS20,
  author       = {Leonardo C. T. Bezerra and
                  Manuel L{\'{o}}pez{-}Ib{\'{a}}{\~{n}}ez and
                  Thomas St{\"{u}}tzle},
  title        = {Automatically Designing State-of-the-Art Multi- and Many-Objective
                  Evolutionary Algorithms},
  journal      = {Evol. Comput.},
  volume       = {28},
  number       = {2},
  pages        = {195--226},
  year         = {2020},
  url          = {https://doi.org/10.1162/evco\_a\_00263},
  doi          = {10.1162/EVCO\_A\_00263},
  bibsource    = {dblp computer science bibliography, https://dblp.org}
}

@article{DBLP:journals/tec/SteinB25,
  author       = {Niki van Stein and
                  Thomas B{\"{a}}ck},
  title        = {LLaMEA: {A} Large Language Model Evolutionary Algorithm for Automatically
                  Generating Metaheuristics},
  journal      = {{IEEE} Trans. Evol. Comput.},
  volume       = {29},
  number       = {2},
  pages        = {331--345},
  year         = {2025},
  url          = {https://doi.org/10.1109/TEVC.2024.3497793},
  doi          = {10.1109/TEVC.2024.3497793},
  bibsource    = {dblp computer science bibliography, https://dblp.org}
}

@article{DBLP:journals/corr/abs-2405-20132,
  author       = {Niki van Stein and
                  Thomas B{\"{a}}ck},
  title        = {LLaMEA: {A} Large Language Model Evolutionary Algorithm for Automatically
                  Generating Metaheuristics},
  journal      = {CoRR},
  volume       = {abs/2405.20132},
  year         = {2024},
  url          = {https://doi.org/10.48550/arXiv.2405.20132},
  doi          = {10.48550/ARXIV.2405.20132},
  eprinttype    = {arXiv},
  eprint       = {2405.20132},
  bibsource    = {dblp computer science bibliography, https://dblp.org}
}

@inproceedings{DBLP:conf/icml/0044TY0LWL024,
  author       = {Fei Liu and
                  Xialiang Tong and
                  Mingxuan Yuan and
                  Xi Lin and
                  Fu Luo and
                  Zhenkun Wang and
                  Zhichao Lu and
                  Qingfu Zhang},
  title        = {Evolution of Heuristics: Towards Efficient Automatic Algorithm Design
                  Using Large Language Model},
  booktitle    = {Forty-first International Conference on Machine Learning, {ICML} 2024,
                  Vienna, Austria, July 21-27, 2024},
  publisher    = {OpenReview.net},
  year         = {2024},
  url          = {https://openreview.net/forum?id=BwAkaxqiLB},
  bibsource    = {dblp computer science bibliography, https://dblp.org}
}

@inproceedings{DBLP:conf/aaai/Yao00L0025,
  author       = {Shunyu Yao and
                  Fei Liu and
                  Xi Lin and
                  Zhichao Lu and
                  Zhenkun Wang and
                  Qingfu Zhang},
  editor       = {Toby Walsh and
                  Julie Shah and
                  Zico Kolter},
  title        = {Multi-Objective Evolution of Heuristic Using Large Language Model},
  booktitle    = {AAAI-25, Sponsored by the Association for the Advancement of Artificial
                  Intelligence, February 25 - March 4, 2025, Philadelphia, PA, {USA}},
  pages        = {27144--27152},
  publisher    = {{AAAI} Press},
  year         = {2025},
  url          = {https://doi.org/10.1609/aaai.v39i25.34922},
  doi          = {10.1609/AAAI.V39I25.34922},
  bibsource    = {dblp computer science bibliography, https://dblp.org}
}

@article{DBLP:journals/nature/RomeraParedesBNBKDREWFKF24,
  author       = {Bernardino Romera{-}Paredes and
                  Mohammadamin Barekatain and
                  Alexander Novikov and
                  Matej Balog and
                  M. Pawan Kumar and
                  Emilien Dupont and
                  Francisco J. R. Ruiz and
                  Jordan S. Ellenberg and
                  Pengming Wang and
                  Omar Fawzi and
                  Pushmeet Kohli and
                  Alhussein Fawzi},
  title        = {Mathematical discoveries from program search with large language models},
  journal      = {Nat.},
  volume       = {625},
  number       = {7995},
  pages        = {468--475},
  year         = {2024},
  url          = {https://doi.org/10.1038/s41586-023-06924-6},
  doi          = {10.1038/S41586-023-06924-6},
  bibsource    = {dblp computer science bibliography, https://dblp.org}
}

@article{DBLP:journals/corr/abs-2506-13131,
  author       = {Alexander Novikov and
                  Ng{\^{a}}n Vu and
                  Marvin Eisenberger and
                  Emilien Dupont and
                  Po{-}Sen Huang and
                  Adam Zsolt Wagner and
                  Sergey Shirobokov and
                  Borislav Kozlovskii and
                  Francisco J. R. Ruiz and
                  Abbas Mehrabian and
                  M. Pawan Kumar and
                  Abigail See and
                  Swarat Chaudhuri and
                  George Holland and
                  Alex Davies and
                  Sebastian Nowozin and
                  Pushmeet Kohli and
                  Matej Balog},
  title        = {AlphaEvolve: {A} coding agent for scientific and algorithmic discovery},
  journal      = {CoRR},
  volume       = {abs/2506.13131},
  year         = {2025},
  url          = {https://doi.org/10.48550/arXiv.2506.13131},
  doi          = {10.48550/ARXIV.2506.13131},
  eprinttype    = {arXiv},
  eprint       = {2506.13131},
  bibsource    = {dblp computer science bibliography, https://dblp.org}
}

@article{DBLP:journals/itor/CamachoVillalonDS23,
  author       = {Christian Leonardo Camacho{-}Villal{\'{o}}n and
                  Marco Dorigo and
                  Thomas St{\"{u}}tzle},
  title        = {Exposing the grey wolf, moth-flame, whale, firefly, bat, and antlion
                  algorithms: six misleading optimization techniques inspired by \emph{bestial}
                  metaphors},
  journal      = {Int. Trans. Oper. Res.},
  volume       = {30},
  number       = {6},
  pages        = {2945--2971},
  year         = {2023},
  url          = {https://doi.org/10.1111/itor.13176},
  doi          = {10.1111/ITOR.13176},
  bibsource    = {dblp computer science bibliography, https://dblp.org}
}

@book{DBLP:series/ncs/2020DN,
  editor       = {Benjamin Doerr and
                  Frank Neumann},
  title        = {Theory of Evolutionary Computation - Recent Developments in Discrete
                  Optimization},
  series       = {Natural Computing Series},
  publisher    = {Springer},
  year         = {2020},
  url          = {https://doi.org/10.1007/978-3-030-29414-4},
  doi          = {10.1007/978-3-030-29414-4},
  isbn         = {978-3-030-29413-7},
  bibsource    = {dblp computer science bibliography, https://dblp.org}
}

@phdthesis{Krasnogor2002_thesis,
  title={Studies on the theory and design space of memetic algorithms},
  author={Natalio Krasnogor},
  year={2002},
  school={University of the West of England},
  address={Bristol}
}

@techreport{Moscato1989_report,
  title={On Evolution, Search, Optimization, Genetic Algorithms and Martial Arts - Towards Memetic Algorithms},
  author={Pablo Moscato},
  institution = "California Institute of Technology",
  year        = "1989",
  type        = "Caltech Con-Current Computation Program 158-79",
  number      = "Technical Report C3P 826",
  address     = "Pasadena",
}

@ARTICLE{Ong2006,
  author={Yew-Soon Ong and Meng-Hiot Lim and Ning Zhu and Kok-Wai Wong},
  journal={IEEE Transactions on Systems, Man, and Cybernetics, Part B (Cybernetics)}, 
  title={Classification of adaptive memetic algorithms: a comparative study}, 
  year={2006},
  volume={36},
  number={1},
  pages={141-152},
  doi={10.1109/TSMCB.2005.856143}
}

@inproceedings{Long2022_gecco,
    author = {Long, Fu Xing and van Stein, Bas and Frenzel, Moritz and Krause, Peter and Gitterle, Markus and B\"{a}ck, Thomas},
    title = {Learning the characteristics of engineering optimization problems with applications in automotive crash},
    year = {2022},
    isbn = {9781450392372},
    publisher = {Association for Computing Machinery},
    address = {New York, NY, USA},
    url = {https://doi.org/10.1145/3512290.3528712},
    doi = {10.1145/3512290.3528712},
    booktitle = {Proceedings of the Genetic and Evolutionary Computation Conference},
    pages = {1227–1236},
    numpages = {10},
    location = {Boston, Massachusetts},
    series = {GECCO '22}
}

@article{Long2025_surrogate,
  author       = {Long, Fu Xing and van Stein, Niki and Frenzel, Moritz and Krause, Peter and Gitterle, Markus and B{\"a}ck, Thomas},
  title        = {Surrogate-based automated hyperparameter optimization for expensive automotive crashworthiness optimization},
  journal      = {Structural \& Multidisciplinary Optimization},
  volume       = {68},
  number       = {4},
  article-number = {68},
  year         = {2025},
  doi          = {10.1007/s00158-025-03989-x},
  url          = {https://doi.org/10.1007/s00158-025-03989-x}
}

@inproceedings{deWinter2024,
  author    = {de Winter, Roy and Long, Fu Xing and Thomaser, Andre and B{\"a}ck, Thomas H. W. and van Stein, Niki and Kononova, Anna V.},
  title     = {Landscape Analysis Based vs. Domain-Specific Optimization for Engineering Design Applications: A Clear Case},
  booktitle = {Proceedings of the 2024 IEEE Conference on Artificial Intelligence (CAI)},
  pages     = {776--781},
  year      = {2024},
  doi       = {10.1109/CAI59869.2024.00148},
  publisher = {IEEE}
}

@article{NFLT,
  author        = "D. Wolpert and W. Macready",
  title         = "No free lunch theorems for optimization",
  journal       = "IEEE Transactions on Evolutionary Computation",
  volume        = "1",
  issue         = "1",
  year          = "1997",
  doi ="10.1109/4235.585893",
  pages         = "67--82"
}

\end{document}